%% file: yj_v06.tex
\documentclass[final]{cvpr}

\usepackage{times}
\usepackage{epsfig}
\usepackage{graphicx}
\usepackage{amsmath}
\usepackage{amssymb}
\usepackage{color}
\usepackage{bbm}
\usepackage{multirow}
\usepackage{algorithm}  
\usepackage{algorithmicx}  
\usepackage{algpseudocode}  
\usepackage{capt-of}
\usepackage{mathtools, cuted}

\usepackage[pagebackref=true,breaklinks=true,colorlinks,bookmarks=false]{hyperref}

\def\cvprPaperID{4293} 
\def\confYear{CVPR 2021}

\begin{document}

\title{Semantic Scene Completion via Integrating Instances and Scene in-the-Loop}
\author{Yingjie Cai\textsuperscript{1}\thanks{The first two authors contribute equally to this work.}, 
\ Xuesong Chen\textsuperscript{1}\footnotemark[1],
\ Chao Zhang\textsuperscript{3}, 
\ Kwan-Yee Lin\textsuperscript{1, }\textsuperscript{2}\thanks{H. Li and K. Lin are the co-corresponding authors.},
\ Xiaogang Wang\textsuperscript{1},
\ Hongsheng Li\textsuperscript{1,}\textsuperscript{4}\footnotemark[2] \\
\textsuperscript{1}CUHK-SenseTime Joint Laboratory, The Chinese University of Hong Kong \\
\textsuperscript{2}SenseTime Research and Tetras.AI \
\textsuperscript{3}Samsung Research Institute China - Beijing (SRC-B) \\
\textsuperscript{4}School of CST, Xidian University\qquad \\
{\tt\small caiyingjie@link.cuhk.edu.hk, cedarchen@pku.edu.cn}
}

\maketitle
\thispagestyle{empty}
\pagestyle{empty}

\begin{strip}
    \centering
    \vspace{-1.5cm}
    \includegraphics[width=1.0\textwidth]{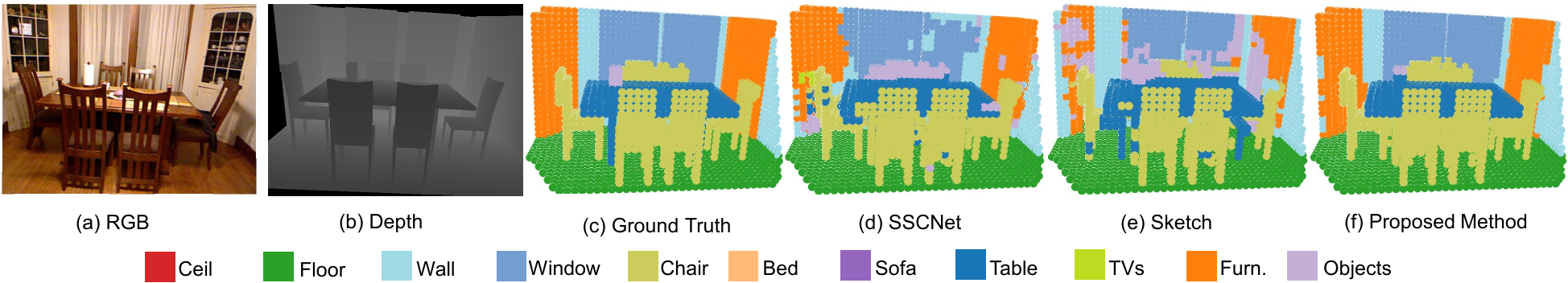}
    \captionof{figure}{\textbf{Visualization of Semantic Scene Completion on NYU Dataset.} From left to right: (a) RGB input, (b) depth map, (c) ground truth of semantic scene completion, (d) results of SSCNet~\cite{song2017semantic-sscnet}, (e) results of Sketch~\cite{Chen_2020_SketchAwareSSC}, (f) results of the proposed method. Our method recovers better shape details like the legs of chairs and generates more reasonable results on easy confusing regions like windows and wall compared with the classic SSC method SSCNet~\cite{song2017semantic-sscnet},  and the state-of-the-art method Sketch~\cite{Chen_2020_SketchAwareSSC}.}
    \label{fig:intro}
\end{strip}
\begin{abstract}
Semantic Scene Completion aims at reconstructing a complete 3D scene with precise voxel-wise semantics from a single-view depth or RGBD image. It is a crucial but challenging problem for indoor scene understanding. In this work, we present a novel framework named Scene-Instance-Scene Network (\textit{SISNet}), which takes advantages of both instance and scene level semantic information.
Our method is capable of inferring fine-grained shape details as well as nearby objects whose semantic categories are easily mixed-up. The key insight is that we decouple the instances from a coarsely completed semantic scene instead of a raw input image to guide the reconstruction of instances and the overall scene. SISNet conducts iterative scene-to-instance (SI) and instance-to-scene (IS) semantic completion. 
Specifically, the SI is able to encode objects' surrounding context for effectively decoupling instances from the scene and each instance could be voxelized into higher resolution to capture finer details. With IS, fine-grained instance information can be integrated back into the 3D scene and thus leads to more accurate semantic scene completion. 
Utilizing such an iterative mechanism, the scene and instance completion benefits each other to achieve higher completion accuracy. Extensively experiments show that our proposed method consistently outperforms state-of-the-art methods on both real NYU, NYUCAD and synthetic SUNCG-RGBD datasets. 
\end{abstract}
\vspace{-0.5cm}
\section{Introduction}
A comprehensive indoor scene understanding in 3D behavior is pivotal for many computer vision tasks, such as robot navigation, virtual/augmented reality and localization, to name a few. 
Semantic Scene Completion (SSC) aims at reconstructing full voxel-wise semantics of a 3D scene from a single-view image.
However, as the real-world scenarios always have various object shapes/sizes, crowded placements, and object-to-object occlusions, precisely reconstructing and understanding the semantics of a whole 3D scene from {partial observations} is of great challenges.

To overcome the incomplete and complex nature of 3D scene understanding, various 3D scene/instance completion and segmentation techniques have been proposed in recent years. Most existing Semantic Scene Completion (SSC) methods~\cite{guedes2018semantic,garbade2019two,liu2018see,li2019rgbd,song2017semantic,dourado2019edgenet,liu2018see,li2019rgbd,zhang2018semantic,guo2018view,garbade2019two,zhang2019cascaded-ccpnet,Li2020aicnet} pass the incomplete scene from view frustum to a neural network. They design complex modules to aggregate multi-level context information to guide the prediction of volumetric occupancy of invisible regions and semantic categories over the whole scene. However, objects' shapes usually cannot be reliably reconstructed in fine-details and the semantic categories of close-by objects are easily mixed-up, due to the limited resolution of voxelization and absence of instance-level constraints, as illustrated by the chairs and windows in Figure~\ref{fig:intro} (e). Some recent Semantic Instance Completion (SIC) methods~\cite{hou2020revealnet,hou2019sis,kundu20183d,avetisyan2020scenecad} attempt to reconstruct certain types of objects in the scene to help the understanding of the 3D scene. They general follow the pipeline of detecting instances and then completing each individuals or aligning with CAD models to achieve fine-grained reconstruction. Instance-level completion can greatly preserve the structures and details of the objects. However, detecting objects from partially observed scenes itself is a challenging problem. In addition, the surrounding semantic information of the scene and more complete shape are helpful for better distinguishing each individuals. Such valuable knowledge is much ignored in the current instance detection and completion methods.

In this paper, we aims to design a mechanism that can efficiently propagate and integrate the information between the scene and instances to achieve more accurate SSC prediction. To this end, we introduce a {Scene-Instance-Scene Network}, a novel framework that conducts {iterative} {\textit {scene-to-instance}} and {\textit{instance-to-scene}} semantic completion. Unlike existing SIC methods that conduct instance completion on the raw partial observations or SSC methods that only consider the scene-level knowledge, our framework takes full advantages of the 3D voxel-wise scene-level ground truths for assisting instance completion, which in turn, improves the whole scene semantic completion in an iterative manner. 

Specifically, the proposed method first aggregates multi-modal information from the visible regions' 2D semantic segmentation maps and truncated signed distance function (TSDF)\footnote{TSDF is a representation to encode depth volume, where every voxel stores the distance value $d$ to its closest surface and the sign of the value indicates whether the voxel is in visible or invisible spaces.} to generate a coarsely completed 3D semantic scene. Given the roughly completed scene, the follow-up scene-to-instance completion localizes each instances and locally voxelizes them into higher resolution to recover detailed 3D shapes, which are then placed back to the scene to further promote the scene's completion. 
The intuition behind this design is that objects are the main components of the scene and they are tightly correlated with the scene. For example, when the windows are well reconstructed, the surrounding walls can be easily inferred. {It is worth noting that we iterate the scene-instance-scene completion in a weight-sharing manner by training with multi-stage data simultaneously.} The iterative completion mechanism enables instances and scene information to be fully propagated and integrated without extra parameters, resulting in more accurate and comprehensive understanding of the 3D scene.

In summary, our contributions are three folds:
\begin{itemize}
    \vspace{-0.25cm}
    \item We introduce a novel framework, Scene-Instance-Scene Network (SISNet), that fully takes advantages of both scene- and instance-level knowledge, to achieve high-quality semantic scene completion. Our method greatly boosts the instance completion performance especially for semantic confusing objects and shape's details with a coarsely completed 3D scene's knowledge in hand. And in turn, fine-detail instance completion leads to more accurate prediction of semantic scene completion.
    \vspace{-0.25cm}
    \item We design a scene-instance-scene iterative completion mechanism, which gradually improves the completion results to better reconstruct the 3D scene without increasing extra parameters.
    \vspace{-0.25cm}
    \item Experimental results demonstrate that the proposed method outperforms state-of-the-art semantic scene completion methods significantly on both synthetic and real datasets.
\end{itemize}
\vspace{-0.2cm}
\section{Related Work}
\vspace{-0.1cm}
\subsection{Semantic Instance Completion}
\vspace{-0.1cm}
Semantic instance completion aims to detect individual instances in a scene and infer their complete object geometry. Most existing methods~\cite{hou2020revealnet,hou2019sis,kundu20183d,avetisyan2020scenecad} attempt to reconstruct objects in the scene by detecting instances and then completing shape with CAD models or completion head. 
Here we briefly review recent works for detection and completion.
For detection, since the irregularity of 3D point data, \cite{song2016deep,zhou2018voxelnet} group points into voxels and use 3D CNN to learn features to predict 3D boxes. However, the their voxelization suffer from information loss. Therefore, \cite{qi2017pointnet,qi2017pointnet++} are proposed to learn the point cloud features directly, which greatly boost the development of point-based 3D detection. \cite{qi2019deep,yang20203dssd,shi2019points} design end-to-end frameworks to detect 3D objects directly in raw point cloud. 
Compared with existing methods, our method utilizes initial completed scene results with point-level semantic information to yield more accurate 3D boxes, especially for semantic confusing objects.
\begin{figure*}[t!]
\centering
\includegraphics[width=\textwidth]{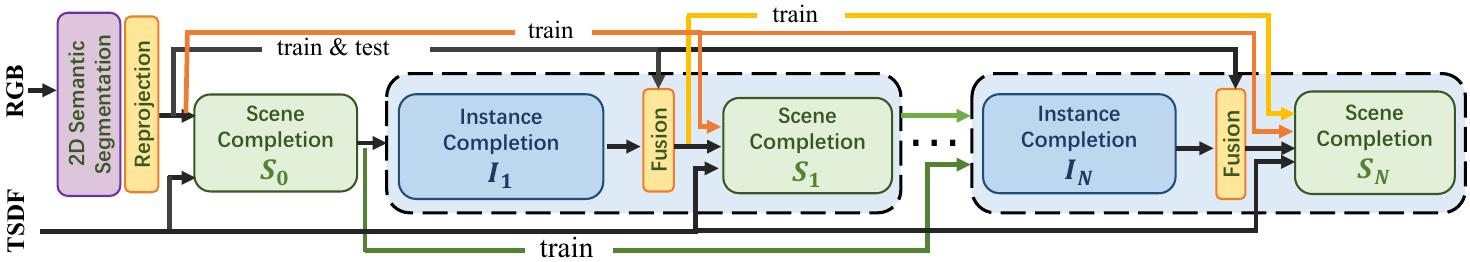}
\vspace{-4mm}
\caption{{\textbf{Overview of the Proposed Method}. SISNet consists of iterative scene-to-instance completion and instance-to-scene completion stages. Given single-view RGBD images, TSDF from the depth map and semantic volume from the reprojection of 2D semantic segmentation are input into the \textit{initial} scene completion $(S_0)$ to obtain a coarse completed scene, which is then fed into the \textit{first} instance completion ($I_1$) to locally recover the instance details. The completed instances are further merged into the semantic volume input to obtain better scene completion prediction ($S_1$). More $I_{i}$-$S_i$ iterations are conducted to promote information integration between the instances and the scene in a weight-sharing manner without extra parameters.}}
\vspace{-0.5cm}
\label{fig:arch}
\end{figure*}
For instance shape completion, \cite{pauly2008discovering,speciale2016symmetry,thrun2005shape} employ reasoning geometric cues, such as symmetries in meshes or point clouds, to complete partial inputs. Alternatively, some methods \cite{nan2012search,shao2012interactive,kim2012acquiring} make up missing geometry by matching with 3D CAD models from a large shape database. \cite{yuan2018pcn,tchapmi2019topnet,huang2020pf,wang2020cascaded,wen2020point} operate on raw partial point clouds directly to reconstruct a completed point cloud through end-to-end network. \cite{xie2020grnet} introduces 3D grids as intermediate representations to regularize point clouds and then employ a 3D CNN network for point cloud completion.
Different from these methods, our instance completion exploits reliable and in hand semantic category priors provided by detection to recover fine-grained details.

\subsection{Semantic Scene Completion}
\vspace{-0.2cm}
Semantic scene completion aims to predict volumetric occupancy and semantic labels simultaneously. SSCNet~\cite{song2017semantic-sscnet} first proposes the combination of completion and segmentation and proves that they can promote each other in an end-to-end network. ForkNet~\cite{wang2019forknet} proposes a multi-branch architecture and utilizes generative models to solve the disadvantages of real scene data. SCFusion~\cite{wu2020scfusion} targets at real-time scene reconstruction from a sequence of depth maps. CCPNet~\cite{zhang2019cascaded-ccpnet} proposes a cascaded context pyramid which progressively restores details of objects and improves the labeling coherence. As complementary information of depth map, RGB images are leveraged by some methods including~\cite{guedes2018semantic,garbade2019two,liu2018see} where 3D CNN is employed to fuse two-stream information, including RGB-based semantic label and depth to improve the semantic discrimination of features. To better fused multi-scale RGB-D features, DDRNet~\cite{li2019rgbd-ddrnet} proposes a light-weight dimensional decomposition residual network. AICNet~\cite{Li2020aicnet} modifies the standard 3D convolution so that kernels with varying sizes. Sketch \cite{Chen_2020_SketchAwareSSC} proposes a 3D sketch-aware feature embedding to explicitly encode geometric information to guide the prediction.
However, our method effectively exploits the relations between instances reconstruction and scene completion in an interactive manner, which effectively recovers the details while guarantees the rigid shapes of objects.
\vspace{-0.2cm}
\section{Methodology}
\vspace{-0.2cm}
\subsection{Architecture Overview}
\vspace{-0.1cm}
The goal of our work is to precisely reconstruct the voxel-wise semantics of a 3D scene from single-view RGB-D images. Since the real-world scenes always contain objects with various sizes/shapes and crowded arrangements, the system must be able to distinguish close-by objects and infer the reasonable shape details of each of them. To this end, we propose the Scene-Instance-Scene Network (SISNet), which consists of a series of in-the-loop scene-to-instance (SI) and instance-to-scene (IS) completion stages. Taking a pair of RGB and depth images as inputs, the SISNet first identifies the visible surfaces by re-projecting each pixel and its 2D semantic label map onto the view frustum, forming two volumes. Then these volumes are fed into the proposed SISNet to produce a completed 3D voxel-wise representation storing volumetric occupancy and semantic labels. The overall architecture is illustrated in Figure~\ref{fig:arch}.


Specifically, we first perform an initial scene completion ($S_0$), to aggregate multi-modal partial information from input volumes and output a roughly completed 3D scene prediction. 
This stage is much ignored by existing semantic instance completion methods and is shown to have crucial impacts to the final completion performance.
Given the roughly completed scene from $S_0$, a scene-to-instance completion stage $S_0 \rightarrow I_1$ is designed to localize each instance and voxlizes them into higher resolution to recover 3D shape details. We then conduct instance-to-scene completion ($I_1 \rightarrow S_1$), where the reconstructed instances with fine details are placed back into the 3D scene to form an enhanced input of next scene-to-instance completion. Utilizing such an in-the-loop scene-instance-scene iteration completion mechanism, we iterate scene-to-instance completion $S_i\rightarrow I_{i+1}$ and instance-to-scene completion $I_{i+1} \rightarrow S_{i+1}$ in a weight-sharing manner until convergence. Such a framework enables the information between the instances and the scene to be effectively propagated and integrated without extra parameters, achieving more accurate reconstruction of the 3D scene. Details are described in the following subsections.


\vspace{-0.1cm}
\subsection{Initial Semantic Scene Completion}
\vspace{-0.1cm}
As the first step of our iterative scene-instance-scene completion framework, we perform an initial semantic scene completion $S_0$ to obtain a roughly completed 3D scene. Such a roughly completed 3D scene leverages the valuable voxel-wise semantic ground truths from SSC datasets so that it can capture more complete context and richer details, which helps the follow-up scene-to-instance completion.
The initial scene completion $S_0$ takes a truncated signed distance function (TSDF) volume $V_T$ and a re-projected semantic volume $V_{S_0}$ as its inputs and outputs a 3D semantic volume, where each voxel is assigned with a semantic label $c_i$, $i \in [1, \cdots, C]$, where $C$ denotes the number of semantic categories.

Specifically, we reconstruct the depth map by re-projecting each pixel onto the view frustum and obtain a $60\times36\times60\times1$  TSDF volume $V_T$, where every voxel stores the signed distance value $d$ to its closest surface.
For the semantic volume $V_{S_0}$, we exploit light-weight 2D semantic segmentation networks following~\cite{guedes2018semantic,garbade2019two,liu2018see,li2019rgbd-ddrnet} to obtain visible surface's semantic labels from the RGB image and then re-project it to the 3D space of $60\times36\times60\times C$, where every voxel records a $C$-class semantic segmentation confidence vector. The combination of $V_T$ and $V_S$ can generally provide rich clues as $V_S$ encodes explicit semantic knowledge compared with the raw RGB images.

\begin{figure}[t!]
    \centering
    \includegraphics[height=2.0cm, width=0.48\textwidth]{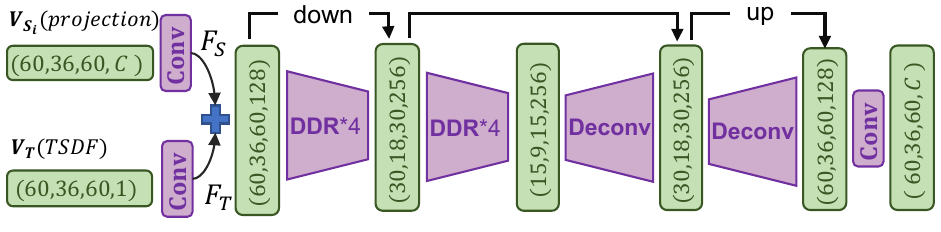}
    \caption{\textbf{Architecture details of initial scene completion.} Given TSDF and semantic volume, to infer missing invisible shape details, we exploit DDR to encode features and propagate information with skip layers, yielding a whole scene completion results.}
    \label{fig:scene}
    \vspace{-0.5cm}
    \end{figure}

Figure~\ref{fig:scene} shows the detailed network structure for initial scene completion $S_0$. We first exploit two convolution blocks, consisting of several convolutions layers followed by ReLU and batch normalization, to transform $V_S$ and $V_T$ into high dimensional features $F_S$ and $F_T$, respectively. Then we feed the summation of them to a 3D Convolution Neural Network (CNN) encoder that includes two blocks with each block consisting of $4$ DDR~\cite{li2019rgbd-ddrnet} modules, which is computation-efficient compared with basic 3D residual block. After encoding, two deconvolution layers are used to upsample the feature embedding back to the original resolution of input and consequently obtain the semantic scene completion predictions after a classification convolution head. In addition, similar to~\cite{chen20203d}, we add several skip connections between encoder and decoder layers for better information and gradient propagation. 

\vspace{-0.1cm}
\subsection{Scene-to-Instance Completion}\label{section:ir}
\vspace{-0.1cm}
\begin{figure*}[t]
    \centering
    \includegraphics[height=3.2cm, width=\textwidth]{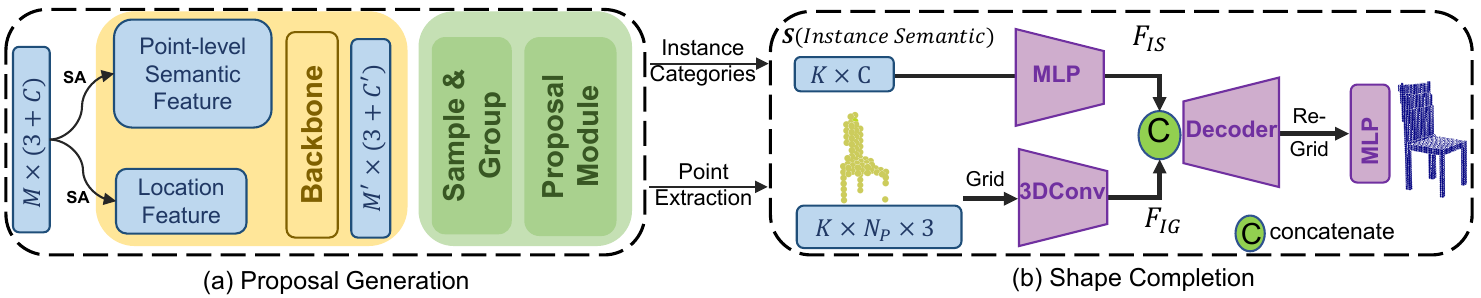}
    \vspace{-0.4cm}
    \caption{\textbf{Architecture details of scene-to-instance completion.} We generate high-quality proposals with a backbone and proposal module to get instances' locations, sizes and categories and then locally voxelizes them into higher resolution to recover detailed 3D shapes.}
    \label{fig:instance}
    \vspace{-0.6cm}
    \end{figure*}
    
Given the roughly completed scene, the scene-to-instance completion aims to localize each instance and recover their detailed 3D shapes, which consists of two components: proposal generation (Figure~\ref{fig:instance} (a)) and instance shape completion (Figure~\ref{fig:instance} (b)), respectively. 

For instance proposal generation, we follow the general design of VoteNet \cite{qi2019deep}. 
$M$ voxels are randomly sampled from the roughly completed 3D scene of $S_0$ with 129,600 ($60\times 36\times 60$) voxels and converted into a point cloud of size $M \times (3+C)$, where $M$ is the number of points, $3$ is the coordinates dimension of the points, and $C$ denotes each point's confidences of the $C$ semantic classes and one extra height channel.
We utilize set abstraction (SA) layers and feature propagation (FP) layers proposed by PointNet++~\cite{qi2017pointnet++} to learn features from the partial instance points.
Specifically, SA layer selects and groups a set of points and then aggregates features of local region points by max pooling. The FP layer consists of several fully connected layers for point feature encoding. 
We employ two SA layers to extract high-dimensional semantic features and location features, respectively. 
The combination of the two types of features are fed into the backbone, which consists of three SA layers and two FP layers, yielding a subset point set of size 
$M^{\prime} \times \left(3+C^{\prime}\right)$. Then we follow VoteNet \cite{qi2019deep} to learn each point's location offset to the center of an instance if the point belongs to a foreground instance. After offset prediction, the center-aware points are grouped into clusters, which are fed into a multi-layer perceptron (MLP) to generate the final proposals.

For instance completion (see Figure~\ref{fig:instance} (b)), during training phase, we select up to $K$ high-quality proposals in one scene whose distances to the center of ground truths boxes are less than a threshold $\sigma$ and objectness confidences are higher than $\beta$. For inference, we use Non-Maximum Suppression (NMS) operation to filter proposals, following \cite{qi2019deep}. Then we pool $N_P$ points from each proposal and normalize them to canonical coordinates and locally voxelizes the partial points into the higher resolution 3D grid of shape $H\times W\times D \times C$, which enables the network to obtain instance-level supervision and prevents the local details of objects from being overwhelmed by the overall scene completion.
The griding and re-griding operations proposed in~\cite{xie2020grnet} are adopted to perform the conversion between point clouds and the 3D grid.

A 3D CNN encoder is followed to learn spatially-aware geometry embedding $F_{IG}$ from the 3D grid representation, which are then concatenated with instance-level semantic features $F_{IS}$ encoded from the instance's class confidences. It restrains the following decoder to reconstruct a complete instance that is consistent with its class's shape prior, obtaining better convergence results.
Then, with the re-griding operation and a MLP, we can obtain the corresponding reconstructed point cloud of size ${N_{R}\times 3}$ with detailed geometry structure, as shown in Figure~\ref{fig:instance} (b).
\vspace{-0.1cm}
\subsection{Instance-to-Scene Completion and Iterative Refinement}
\vspace{-0.1cm}
With the reconstructed instances at hand, we can project them back into the 3D semantic scene volume to assist the completion of the overall 3D scene. Thus, we voxelize the completed instances so that they can match the resolution of the 3D semantic scene volume. 
which leads to an enhanced instance input for subsequent stage. 

Specifically, we update the semantic category vector of the $V_{S_0}$ volume to obtain $V_{S_1}$ by replacing each instance's voxel labels with the reconstructed labels from $I_1$. Then, the following scene completion stage $S_1$, which shares the same architecture of $S_0$, can enjoy more detailed geometry structure information of partial instances than $S_0$. Moreover, under the supervision of scene's semantic and occupied ground truth, the $S_1$ can further refine both reconstructed instances from the previous $I_1$ stage as well as the background in the overall 3D scene layout, resulting in a more comprehensive understanding of the scene.

To further and fully integrate the instance-level and scene-level information, we propose a weight-sharing iteration framework. The same scene-to-instance completion ($S_i\rightarrow I_{i+1}$) and instance-to-scene completion ($I_{i+1}\rightarrow S_{i+1}$) stages can be iterated to fully integrate information between the instances and the scene for better completion accuracy.
During training, to deal with the different distributions at the multiple iterations, the earlier iterations would generate additional training data for the later stages. Specifically, $S_i$ would collect all the enhanced scene volumes from $I_{i}, I_{i-1}, \dots, I_{1}$ as training data. $I_i$ would utilize all completed scene volumes from $S_{i-1}, S_{i-2}, \dots, S_0$ to provide training data for instance completion. In this way, the trained instance-to-scene completion and scene-to-instance completion stages can well handle the data variations across different iterations.
During inference, 
the trained $S_N$ and $I_N$ would be iteratively and alternatively connected to gradually complete the whole 3D semantic scene.

\subsection{Loss Function}
\vspace{-0.1cm}
For each stage in the method, we adopt the respective loss functions to perform instance and scene completion. For instance completion $I_i$, the loss function consists of two terms for proposals generation and instance completion.

\noindent \textbf{Loss Function for Proposals Generation.~}
The predicted location offset $\Delta x_{i}$ can be supervised by
a regression loss
\begin{equation}
\label{eq1}
\setlength\abovedisplayskip{3pt}
\setlength\belowdisplayskip{3pt}
L_{\text {loc-reg }}=\frac{1}{N_\mathrm{total}} \sum_{i {\rm~on ~objects}}\left\|\Delta x_{i}-\Delta x_{i}^{*}\right\|,
\end{equation}
where ${N_\mathrm{total}}$ is the total number of instances' points, $\Delta x_{i}^{*}$ is the ground truth offset from the point $x_{i}$ to its corresponding instance's center.
To supervise the generation of proposals, we define positive and negative proposals generated from grouped clusters by determining whether these locations' distance to a ground truth object center is within $0.3$ meters or more than $0.6$ meters.

The positive proposals are supervised by a box loss $L_{\mathrm{box}}$ and semantic classification loss $L_{\mathrm{sem-cls}}$.
We use the standard cross entropy loss for $L_{\mathrm{sem-cls}}$ and, following~\cite{qi2019deep,qi2018frustum}, we decouple $L_{\mathrm{box}}$ into center regression, size estimation which use a hybrid of classification and regression formulations. The cross entropy loss is used for size classification and the robust L1-smooth loss is employed for regression:
\begin{align}
\small
\setlength\abovedisplayskip{3pt}
\setlength\belowdisplayskip{3pt}
L_{\mathrm{box}} & =L_{\mathrm {center-reg}}+ \lambda_{1} L_{\mathrm {size-cls}}+L_{\mathrm {size-reg}},
\normalsize
\end{align}
where $\lambda_1$ is a relative weight for the size classification term.
To supervise objectness of proposals, we employ a normalized cross entropy loss $L_{\mathrm {obj-cls}}$, which is performed on both positive and negative samples. In summary, the loss for proposal generation can be written as
\begin{align}
\small
\label{eq1}
\setlength\abovedisplayskip{3pt}
\setlength\belowdisplayskip{3pt}
 L_{\mathrm {det}} & =L_{\mathrm {loc-reg }} + L_{\mathrm {box }}+\lambda_{2} L_{\mathrm {obj-cls }}+\lambda_{3} L_{\mathrm {sem-cls}},
 \normalsize \nonumber
\end{align}
where $\lambda_2$ and $\lambda_3$ are the relative weights for the loss terms.

\noindent \textbf{Loss Function for Instance Completion.~}
To make the reconstructed points capture detailed geometry structures, we exploit the Chamfer Distance (CD) to supervise the reconstruction process. 
Let ${\mathcal{T}=\left\{\left(x_{i}, y_{i}, z_{i}\right)\right\}_{i=1}^{n_{\mathcal{T}}}}$ be the ground truth and ${\mathcal{R}=\left\{\left(x_{i}, y_{i}, z_{i}\right)\right\}_{i=1}^{n_{\mathcal{R}}}}$ be the reconstructed instance point set, where ${n_{\mathcal{T}}}$ and ${n_{\mathcal{R}}}$ denote the numbers of points of ${\mathcal{T}}$ and ${\mathcal{R}}$. The CD loss can be written as
\begin{equation}
\small
\setlength\abovedisplayskip{0.5pt}
\setlength\belowdisplayskip{0.5pt}
\label{eq3}
\begin{split}
L_{\mathrm {CD}}=\frac{1}{n_{\mathcal{T}}} \sum_{t \in \mathcal{T}} \min _{r \in \mathcal{R}}\|t-r\|_{2}^{2}+\frac{1}{n_{\mathcal{R}}} \sum_{r \in \mathcal{R}} \min _{t \in \mathcal{T}}\|t-r\|_{2}^{2}.
\end{split}
\end{equation}
Finally, through a combination of the proposal loss and the CD loss, we can jointly train our instance completion stages, which can be expressed as
\begin{equation}
\setlength\abovedisplayskip{8pt}
\setlength\belowdisplayskip{8pt}
\small
L_\mathrm{instance} = L_\mathrm{det} + L_\mathrm{CD}.
\label{totalloss}
\end{equation}

\noindent \textbf{Loss Function for Scene Completion.~}
The scene completion is supervised by  voxel-wise cross-entropy loss:
\begin{equation}
\setlength\abovedisplayskip{2pt}
\setlength\belowdisplayskip{2pt}
L_\mathrm{scene} = \mathcal \ell_\mathrm{ce} \left( S_{i}(V_T,V_{S_i}), G\right), i \in\left\{0, \cdots, N\right\}
\end{equation}
where $G$ is the ground truth semantic label and $\ell_\mathrm{ce}$ denotes the multi-class cross-entropy loss.

\input{tables/resNYU.tex}
\input{tables/resNYUCAD.tex}
\input{tables/resSUNCG.tex}

\vspace{-0.2cm}
\section{Experiments}
\vspace{-0.1cm}
\subsection{Datasets and Evaluation Metrics}
\noindent \textbf{Datasets.} We evaluate the proposed method on both real and synthetic datasets. The two real datasets are the popular NYU Depth V2 \cite{nyudv2} (denoted as NYU in the following) and NYUCAD~\cite{firman2016structured}. They both consists of 1449 indoor scenes and there are 795 for training and 654 for test. We follow \cite{song2017semantic-sscnet} and use the 3D annotated labels provided by \cite{rock2015completing} for semantic scene completion task. 

\textbf{NYU} dataset is a challenging dataset, as its depth measurements have errors compared with ground-truths. To address the misalignment of some depth maps and their corresponding label volumes. \textbf{NYUCAD} uses the depth maps generated from the projections of the 3D annotations, however, the RGB image and depth map are not aligned. 
\textbf{SUNCG-RGBD}~\cite{liu2018see-satnet} is a subset dataset of SUNCG~\cite{song2017semantic-sscnet}, consisting of 13,011 training samples and 499 testing samples and all samples provide depth images and corresponding RGB images which are not provided in SUNCG.

\noindent \textbf{Evaluation Metrics.} We follow SSCNet \cite{song2017semantic-sscnet} which evaluates semantic scene completion with two types of metrics. The first one focuses on evaluating semantic performance $i.e.$, semantic scene completion (SSC) and the other one cares more about scene completion (SC), which only measures the accuracy of occupancy instead of all categories. For SSC, we evaluate the IoU of each category on both surface and occluded voxels in the view frustum. For SC, we treat all voxels as binary predictions, $i.e.$, free or non-free. We use recall, precision and voxel-wise intersection over union (IoU) as evaluation metrics to evaluate the binary IoU on occluded voxels in the view frustum.

\begin{figure*}[t!]
\centering
\includegraphics[height=7.6cm, width=\textwidth]{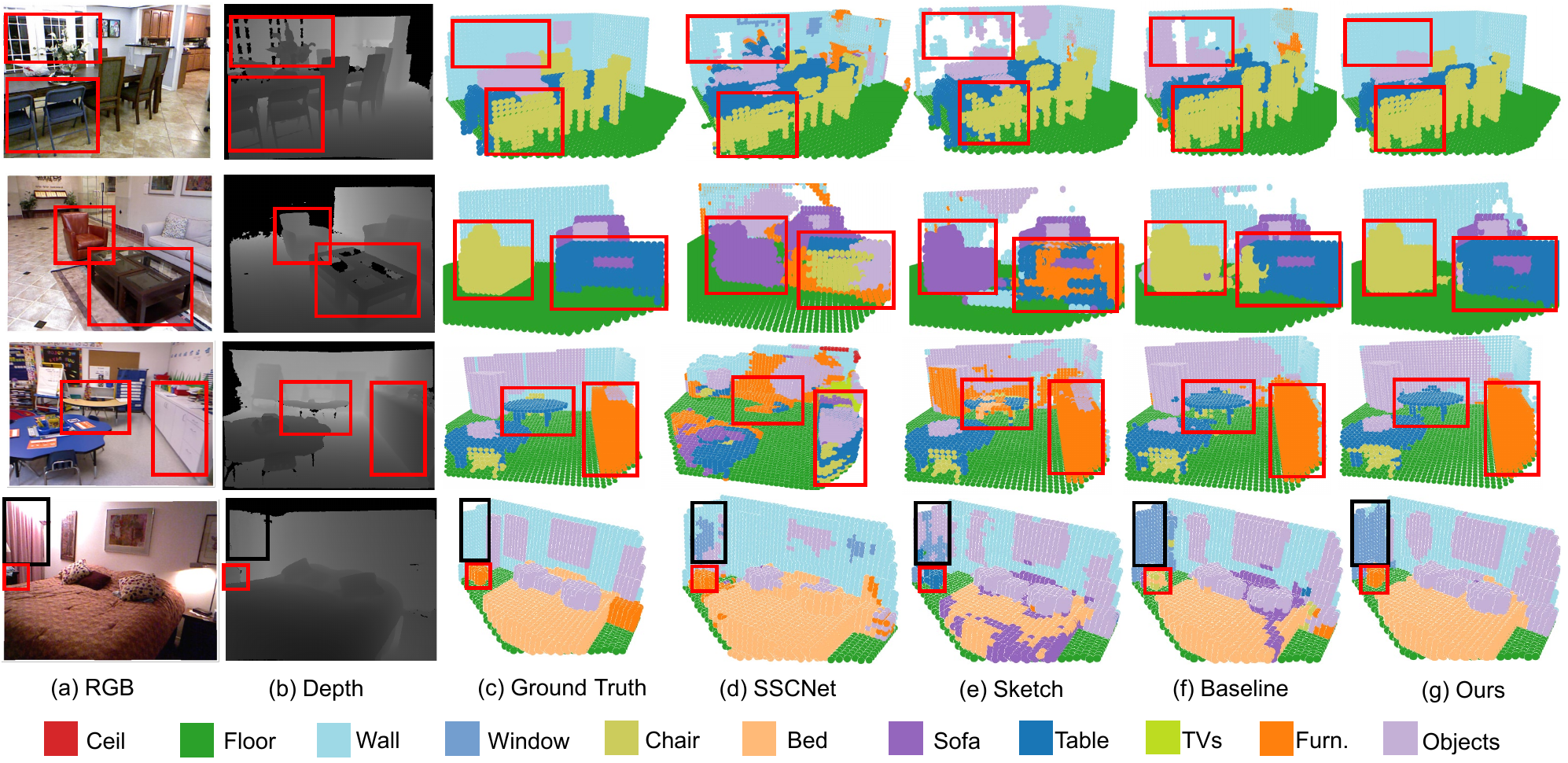}
\caption{\textbf{Semantic Scene Completion results on NYU dataset.} From left to right: (a) RGB input, (b) Depth, (c) ground truth, (d) results of SSCNet~\cite{song2017semantic-sscnet}, (e) results of Sketch~\cite{Chen_2020_SketchAwareSSC}, (f) baseline (without using instance completion), (g) our results. Our results achieve higher voxel-level accuracy compared with SSCNet~\cite{song2017semantic-sscnet} and Sketch~\cite{Chen_2020_SketchAwareSSC}. Better viewed in color and zoom in.}
\vspace{-0.5cm}
\label{fig:nyu-res}
\end{figure*}
\vspace{-0.2cm}
\subsection{Implementation Details}
\vspace{-0.2cm}
We use PyTorch framework to implement our experiments with $8$ NVIDIA 1080 Ti GPUs. Specifically, for instance completion, we adopt an Adam optimizer with an initial learning rate (lr) $10^{-3}$ and $10$ batch size per GPU to train detection backbone with $180$ epochs. The lr is decreased by 10$\times$ at iterations 80, 120 and 160. Then, loading the pre-trained detection weights, we jointly train the detection backbone with $10^{-4}$ lr and instance shape completion with $10^{-3}$ lr for another 20 epochs. 
For scene completion, we employ an SGD optimizer with initial lr $0.1$, batch size $2$ per GPU 
to train NYU and SUNCG-RGBD with $250$ and $160$ epochs, respectively. We adopt a poly learning rate policy where the lr is changed with the iteration number $t$ as $(1-\frac{iter}{max\_iter})^{0.9}$. Our final model employs 2 iterations ($N$ = $2$) to obtain the final results.
We use $\lambda_{1}$ = $\lambda_{3}$ = $0.1$, $\lambda_{2}$ = $0.5$ for the combination of losses. 
\vspace{-0.2cm}
\subsection{Comparisons with State-of-the-art Methods}
\vspace{-0.1cm}
We conduct experiments on NYU, NYUCAD and SUNCG-RGBD datasets, which demonstrates the superiority of our method to the existing  state-of-the-art methods.


\input{tables/ablation-iteration.tex}
\input{tables/ablation-cost.tex}
\input{tables/ablation-instance.tex}
\input{tables/ablation-order.tex}
\noindent \textbf{Comparison on NYU and NYUCAD datasets.~}
Tables~\ref{tab:SotaOnNYU} and~\ref{tab:SotaOnNYUCAD} show the performances of different methods on the NYU and NYUCAD datasets. We can see that our SISNet achieves the best performance on both datasets.
For the SC metric, the proposed method outperforms Sketch~\cite{chen20203d} by $6.5\%$ and $2.3\%$ on NYU and NYUCAD, respectively. For the SSC metric, our approach achieves $49.8\%$ and $59.9\%$, and surpasses Sketch~\cite{chen20203d} by $8.7\%$ and $4.7\%$, respectively.

Our approach consistently outperforms state-of-the-art methods with comparable parameters as Sketch \cite{Chen_2020_SketchAwareSSC} (when using BiSeNet~\cite{yu2018bisenet}). 
When using our designed DeepLabv3~\cite{chen2017deeplab} based method, 
as shown in Tables~\ref{tab:SotaOnNYU},~\ref{tab:SotaOnNYUCAD} and~\ref{tab:ablation-cost}, the improvements are even greater, achieving $11.3\%$ and $8.3\%$ in terms of SSC on NYU and NYUCAD, respectively. Also note that our SISNet's output resolution (60, 36, 60) is on par or lower than those of existing methods, denoting that our voxel-wise accuracy is much higher.
Our design that locally increases the resolution of instance in the intermediate process of the network can achieves better results and dose not increase computation cost significantly.
In addition, we observe that our instance-classes' completion performances consistently surpass existing methods like TVs, whose gains can reach up to 10\%. We attribute the improvement to the exploitation of the completed 3D scene semantics, which greatly boost the performance of instance completion with well preserved shape details.

\noindent \textbf{Comparison on SUNCG-RGBD Dataset.~}
We also conduct experiments on the SUNCG-RGBD~\cite{liu2018see-satnet} dataset to validate the generalization of the proposed method on large-scale dataset. As shown in Table~\ref{tab:SotaOnSUNCG}, with the same input and comparable computation cost, our method outperforms Sketch with  considerable margins about $8.1\%$ and $23.2\%$ in terms of the overall IoUs of SC and SSC, respectively.

\noindent \textbf{Qualitative Results.~}
Figure~\ref{fig:nyu-res} illustrates the qualitative results on NYU. More results please refer to supplementary materials. Although the previous methods work well for some scenes, they usually fail to deal with complex regions consisting of semantic confusing objects or instances with complex structures. By leveraging the complete 3D scene layout and semantics, the instance and scene level information can be effectively propagated and integrated.

\vspace{-0.2cm}
\subsection{Ablation Study}
\vspace{-0.1cm}
To verify the effects of each component of our SISNet, we employ the initial scene completion, $i.e.$ $S_0$, as our baseline and perform ablation studies on the three datasets.

\noindent \textbf{Effects of instance completion.~}To evaluate the benefits of introducing instance completion to SSC, we compare an alternative strategy that skips the instance completion and directly refine the roughly estimated results of $S_0$ with another scene completion $S_1$. Table~\ref{tab:ablation-instance} shows the quantitative results of the compared scheme, denoted as \textit{+ $S_1$ only} and \textit{+Iter $I$-$S$}. Specifically, the SC drop $1.0\%$, $1.8\%$ and $0.6\%$ on NYU, NYUCAD and SUNCG-RGBD for \textit{+$S_1$ only} scheme. Meanwhile, the SSC results decrease $2.8\%$, $3.8\%$ and $2.1\%$ on the dataset. When the instances are refined again without extra parameters, there will be more obvious gains (\ie, \textit{+2-Iter $I$-$S$}), which proves that the instance completion significantly boosts the scene prediction. 


\noindent \textbf{Effects of scene-to-instance completion.~}We design a comparison experiment that directly performs instance completion without the initial scene completion (denoted by \textit{$I_1$-$S_1$}) to compare with our solution (denoted by \textit{$S_0$-$I_1$-$S_1$}). We compare both instance-level detection accuracy and shape completion performance. The evaluation metrics of detection are mean average precision (mAP) and recall with 3D IoU threshold 0.25 following ~\cite{song2015sun}. The evaluation of instance shape completion uses the SSC metric and only considers instance-classes (excluding ceil, floor, and wall).

Table~\ref{tab:ablation-order} shows results of the above two setups on the three datasets. 
Without taking advantages of the roughly estimated 3D scene, \textit{$I_1$-$S_1$} shows much worse performance on instance detection and shape completion, decreasing about $10\%$ (as shown by the columns 3-5 of Table~\ref{tab:ablation-order}). In contrast, our proposed \textit{$S_0$-$I_1$-$S_1$} shows significant improvements on both detection and instance shape completion. It shows about $1\%$, $2\%$ and $1\%$ gains on semantic scene completion with NYU, NYUCAD and SUNCG-RGBD, respectively. It demonstrates that the initial scene completion greatly assists the follow-up instance detection and completion by fully utilizing the valuable scene completion ground truths.


\noindent \textbf{Effects of scene-instance-scene iterations.~}
To evaluate the effect of our proposed scene-instance-scene iteration mechanism, we stack different number of such iterations in a weight-sharing manner (denoted as +\textit{2-Iter I-S} vs. +\textit{Iter I-S}). As illustrated in Table~\ref{tab:ablation-iteration}, based on the rough results of $S_0$, one iteration can improve baseline SSC by $5.2\%$, $4.4\%$ and $3.6\%$ on NYU, NYUCAD and SUNCG-RGBD, respectively. Two iterations can consistently bring about $2\%$ further improvement in terms of SSC. This demonstrates the effectiveness of our iteration mechanism. We also test using more iterations ($N>=3$) and the completion performance does not show obvious improvement ($\leq$1\%). We therefore use $N=2$ iterations for our final model.
\vspace{-0.3cm}
\section{Conclusion}
\vspace{-0.2cm}
In this work, we present a novel framework that decouples the instances from a coarsely completed scene and exploits category priors to guide the reconstruction of the find-grained shape details as well as nearby objects whose semantic categories are easily mixed-up. Furthermore, we design an iterative mechanism, where the scene and instance completion benefits each other. Extensively experiments prove that the proposed method consistently achieves state-of-the-art performance on public benchmark.

\noindent \textbf{Acknowledgments:~}This work is supported in part by Centre for Perceptual and Interactive Intelligence Limited, in part by the General Research Fund through the Research Grants Council of Hong Kong under Grants (Nos. 14208417, 14207319, 14202217, 14203118, 14208619), in part by Research Impact Fund Grant No. R5001-18, in part by CUHK Strategic Fund.
\clearpage

{\small
\bibliographystyle{ieee_fullname}
\bibliography{egbib}
}

\end{document}

%% file: tables/resNYU.tex
\begin{table*}[ht]
\begin{center}
\resizebox{\textwidth}{!}{
\begin{tabular}{|l| c |c c c|c c c c c c c c c c c | c|} 
\hline
 & & \multicolumn{3}{c|}{Scene Completion (SC)} & \multicolumn{12}{c|}{Semantic Scene Completion (SSC)} \\ \hline
Methods  & Resolution & prec. & recall & IoU & ceil. & floor & wall & win. & chair & bed & sofa & table & tvs & furn. & objs. & avg. \\ 
\hline
SSCNet~\cite{song2017semantic-sscnet} & $(240, 60)$ & 57.0 & \textbf{94.5} & 55.1 & 15.1 & 94.7 & 24.4 &  0.0 & 12.6 & 32.1 & 35.0 & 13.0 &  7.8 & 27.1 & 10.1 & 24.7\\
ESSCNet~\cite{zhang2018efficient-esscnet} & $(240, 60)$ & 71.9 & 71.9 & 56.2 & 17.5 & 75.4 & 25.8 &  6.7 & 15.3 & 53.8 & 42.4 & 11.2 &    0 & 33.4 & 11.8 & 26.7\\ 
DDRNet~\cite{li2019rgbd-ddrnet} & $(60, 60)$ & 71.5  & 80.8 & 61.0 & 21.1 & 92.2 & 33.5 & 6.8 & 14.8 & 48.3 & 42.3 & 13.2 & 13.9 & 35.3 &  13.2 & 30.4\\ 
VVNetR-120~\cite{guo2018view-vvnet}  & $(120, 60)$ & 69.8 & 83.1 & 61.1 & 19.3 & 94.8 & 28.0 & 12.2 & 19.6 & 57.0 & 50.5 & 17.6 & 11.9 & 35.6 & 15.3 & 32.9  \\
AICNet~\cite{Li2020aicnet} &$(60, 60)$& 62.4 & 91.8 & 59.2 & 23.2 & 90.8 &32.3& 14.8& 18.2& 51.1& 44.8& 15.2& 22.4& 38.3& 15.7& 33.3\\
TS3D~\cite{garbade2018two-ts3d} & $(240, 60)$ &  - & - & 60.0 & 9.7 & 93.4 & 25.5 & 21.0 & 17.4 & 55.9 & 49.2 & 17.0 & 27.5 & 39.4 & 19.3 & 34.1\\
SATNet~\cite{liu2018see-satnet} & $(60, 60)$ & 67.3 & 85.8 & 60.6 & 17.3 & 92.1 & 28.0 & 16.6 & 19.3 & 57.5 & 53.8 & 17.2 & 18.5 & 38.4 & 18.9 & 34.4 \\
ForkNet~\cite{wang2019forknet}  & $(80, 80)$ & - & - & 63.4 & 36.2 & 93.8 & 29.2 & 18.9 & 17.7 & 61.6 & 52.9 & 23.3 & 19.5 & 45.4 & 20.0 & 37.1 \\
CCPNet~\cite{zhang2019cascaded-ccpnet} & $(240, 240)$ & 74.2  & 90.8 & 63.5 & 23.5 & \textbf{96.3} & 35.7 & 20.2 & 25.8 & 61.4 & 56.1 & 18.1 & 28.1 & 37.8 & 20.1 & 38.5\\ 
Sketch~\cite{Chen_2020_SketchAwareSSC} & $(60, 60)$ & 85.0 & 81.6 & 71.3 & 43.1 & 93.6 & 40.5 & 24.3 & 30.0 & 57.1 & 49.3 & 29.2 & 14.3 & 42.5 & 28.6 & 41.1 \\ 
\hline
\hline
Baseline (BiSeNet) &$(60, 60)$ & 87.6&	78.9& 71.0&
46.9&	93.3&	41.3&	26.7&	30.8&	58.4&	49.5&	27.2&	22.1&	42.2&	28.7&	42.5\\
Ours (BiSeNet) & $(60, 60)$ & 	
\textbf{90.7}&	84.6&	\textbf{77.8}& \textbf{53.9}&	93.2&	\textbf{51.3}&	\textbf{38.0}&	\textbf{38.7}&	\textbf{65.0}&	\textbf{56.3}&	\textbf{37.8}&	\textbf{25.9}&	\textbf{51.3}&	\textbf{36.0}&	\textbf{49.8}\\ 
\hline
Baseline (DeepLabv3) &$(60, 60)$ &88.7&	77.7& 70.8&	46.8&	93.4&	42.0&	32.4&	36.0&	61.1&	55.8&	28.2&	27.6&	45.7&	32.9&	45.6\\
Ours (DeepLabv3) & $(60, 60)$ & 	
\textcolor{blue}{\textbf{92.1}}&	\textcolor{blue}{83.8} &	\textcolor{blue}{\textbf{78.2}} & \textcolor{blue}{\textbf{54.7}} &	\textcolor{blue}{93.8} &	\textcolor{blue}{\textbf{53.2}}&	\textcolor{blue}{\textbf{41.9}}&	\textcolor{blue}{\textbf{43.6}}&	\textcolor{blue}{\textbf{66.2}}&	\textcolor{blue}{\textbf{61.4}}&	\textcolor{blue}{\textbf{38.1}}&	\textcolor{blue}{\textbf{29.8}}&	\textcolor{blue}{\textbf{53.9}}&	\textcolor{blue}{\textbf{40.3}}&	\textcolor{blue}{\textbf{52.4}} \\ 
\hline
\end{tabular}
}
\end{center}
\vspace{-0.2cm}
\caption{\textbf{Results on NYU dataset}. Bold numbers represent the best scores. \textit{(a, b)} means the input and output resolution.}
\label{tab:SotaOnNYU}
\vspace{-0.3cm}
\end{table*}

%% file: tables/resNYUCAD.tex
\begin{table*}[ht]
\begin{center}
\resizebox{\textwidth}{!}{
\begin{tabular}{|l| c |c c c|c c c c c c c c c c c | c|} 
\hline
 & & \multicolumn{3}{c|}{Scene Completion (SC)} & \multicolumn{12}{c|}{Semantic Scene Completion (SSC)} \\ \hline
Methods  & Resolution & prec. & recall & IoU & ceil. & floor & wall & win. & chair & bed & sofa & table & tvs & furn. & objs. & avg. \\ 
\hline
SSCNet~\cite{song2017semantic-sscnet} & $(240, 60)$ & 75.4 & \textbf{96.3} & 73.2 & 32.5 & 92.6 & 40.2 &  8.9 & 33.9 & 57.0 & 59.5 & 28.3 &  8.1 & 44.8 & 25.1 & 40.0\\ 
DDRNet~\cite{li2019rgbd-ddrnet} & $(60, 60)$	& 88.7 & 88.5 & 79.4 & 54.1 & 91.5 & 56.4 & 14.9 & 37.0 & 55.7 & 51.0 & 28.8 & 9.2 & 44.1 & 27.8 & 42.8 \\
AICNet~\cite{Li2020aicnet} &$(60, 60)$& 88.2& 90.3& 80.5& 53.0 &91.2 &57.2& 20.2& 44.6& 58.4& 56.2& 36.2& 9.7& 47.1& 30.4& 45.8\\
TS3D~\cite{garbade2018two-ts3d} & $(240, 60)$   & - & - & 76.1 & 25.9 & 93.8 & 48.9 & 33.4 & 31.2 & 66.1 & 56.4 & 31.6 & 38.5 & 51.4 & 30.8 & 46.2 \\ 
CCPNet~\cite{zhang2019cascaded-ccpnet} & $(240, 240)$ & 91.3 & 92.6 & 82.4 & 56.2 & \textbf{94.6} & 58.7 & 35.1 & 44.8 & 68.6 & 65.3 & 37.6 & 35.5 & 53.1 & 35.2 & 53.2 \\
Sketch~\cite{Chen_2020_SketchAwareSSC} & $(60, 60)$ & 90.6 & 92.2 & 84.2 & 59.7 & 94.3 & 64.3 & 32.6 & 51.7 & 72.0 & \textbf{68.7} & 45.9 & 19.0 & 60.5 & 38.5 &  55.2 \\ 
\hline
\hline
Baseline (BiSeNet) & $(60, 60)$ & 
92.3&	89.0&   82.8 &61.5&	94.2&	62.7&	38.0&	48.1&	69.5&	59.3&	40.1&	25.8&	54.6&	35.3&	53.6\\
Ours (BiSeNet) & $(60, 60)$ & 
\textbf{94.2}&	91.3&	\textbf{86.5}&	\textbf{65.6}&	94.4&	\textbf{67.1}&	\textbf{45.2}&	\textbf{57.2}&	\textbf{75.5}&	66.4&	\textbf{50.9}&	\textbf{31.1}&	\textbf{62.5}&	\textbf{42.9}&	\textbf{59.9}\\
\hline
Baseline (DeepLabv3) & $(60, 60)$ & 92.0&	89.3&	82.8 &62.0&	94.1&	63.3&	43.5&	50.8&	73.3&	63.5&	42.2&	40.6&	58.2&	39.7&	57.4\\
Ours (DeepLabv3) & $(60, 60)$ & 
\textcolor{blue}{\textbf{94.1}}& \textcolor{blue}{91.2}& \textcolor{blue}{\textbf{86.3}} & \textcolor{blue}{\textbf{63.4}}& \textcolor{blue}{94.4}& \textcolor{blue}{\textbf{67.2}}& \textcolor{blue}{\textbf{52.4}}& \textcolor{blue}{\textbf{59.2}}& \textcolor{blue}{\textbf{77.9}}& \textcolor{blue}{\textbf{71.1}}& \textcolor{blue}{\textbf{51.8}}& \textcolor{blue}{\textbf{46.2}}& \textcolor{blue}{\textbf{65.8}}& \textcolor{blue}{\textbf{48.8}}& \textcolor{blue}{\textbf{63.5}}\\
\hline
\end{tabular}
}
\end{center}
\vspace{-0.2cm}
\caption{\textbf{Results on NYUCAD dataset}. Bold numbers represent the best scores. \textit{(a, b)} means the input and output resolution.}
\label{tab:SotaOnNYUCAD}
\vspace{-0.3cm}
\end{table*}

%% file: tables/resSUNCG.tex
\begin{table*}[t!]
\begin{center}
\resizebox{\textwidth}{!}{
\begin{tabular}{|l| c |c c c|c c c c c c c c c c c | c|} 
\hline
  & & \multicolumn{3}{c|}{Scene Completion (SC)} & \multicolumn{12}{c|}{Semantic Scene Completion (SSC)} \\ \hline
Methods  & Resolution 
& prec. & recall & IoU & ceil. & floor & wall & win. & chair & bed & sofa & table & tvs & furn. & objs. & avg. \\ 
\hline

SATNet~\cite{liu2018see-satnet}& $(60, 60)$
& -- & -- & -- & 60.6 & 57.3 & 53.2 & 52.7 & 27.4 & 46.8 & 53.3 & 28.6 & 41.1 & 44.1 & 29.0 & 44.9\\
Sketch~\cite{Chen_2020_SketchAwareSSC}& $(60, 60)$ 
& 	\textbf{94.1}&	86.2& 81.8& 77.9&	82.3&	68.4&	57.9&	35.7&	71.8&	63.7&	45.1&	12.8&	64.2&	32.0&	55.6\\
\hline
\hline
Baseline (BiSeNet) & $(60, 60)$ & 90.6&	94.5& 86.0&	71.0&	86.6&	78.4&	78.7&	53.2&	77.4&	77.3&	60.6&	76.4&	83.7&	60.0&	73.0\\


Ours (BiSeNet) & $(60, 60)$   &
93.3&	\textbf{96.1}&	\textbf{89.9}&	\textbf{85.2}&	\textbf{90.0}&	\textbf{83.7}&	\textbf{80.8}&	\textbf{60.0}&	\textbf{83.5}&	\textbf{80.9}&	\textbf{68.6}&	\textbf{77.3}&	\textbf{86.7}&	\textbf{70.1}&	\textbf{78.8}\\
\hline
Baseline (DeepLabv3) & $(60, 60)$ & 90.1&	94.5&85.6&70.5&	86.2&	77.5&	79.0&	55.5&	82.8&	78.8&	62.4&	69.5&	81.7&	57.8&	72.9\\
Ours (DeepLabv3) & $(60, 60)$ &\textcolor{blue}{92.6} &\textcolor{blue}{\textbf{96.3}} &\textcolor{blue}{\textbf{89.3}}&\textcolor{blue}{\textbf{85.4}} &\textcolor{blue}{\textbf{90.6}} &\textcolor{blue}{\textbf{82.6}} &\textcolor{blue}{\textbf{80.9}} &\textcolor{blue}{\textbf{62.9}} &\textcolor{blue}{\textbf{84.5}} & \textcolor{blue}{\textbf{82.6}}&\textcolor{blue}{\textbf{71.6}} &\textcolor{blue}{\textbf{72.6}} &\textcolor{blue}{\textbf{85.6}}&\textcolor{blue}{\textbf{69.7}} & \textcolor{blue}{\textbf{79.0}}\\
\hline
\end{tabular}
}
\end{center}
\vspace{-0.2cm}
\caption{\textbf{Results on SUNCG-RGBD dataset}. Bold numbers represent the best scores. \textit{(a, b)} means the input and output resolution.}
\label{tab:SotaOnSUNCG}
\vspace{-0.6cm}
\end{table*}

%% file: tables/ablation-iteration.tex
\begin{table*}[t!]
\begin{center}
\resizebox{\textwidth}{!}{
\begin{tabular}{|l|c |c c c|c c c c c c c c c c c | c|} 
\hline
  & & \multicolumn{3}{c|}{Scene Completion (SC)} & \multicolumn{12}{c|}{Semantic Scene Completion (SSC)} \\ \hline
Methods & Trained on & prec. & recall & IoU & ceil. & floor & wall & win. & chair & bed & sofa & table & tvs & furn. & objs. & avg. \\ 
\hline
Baseline & NYU  &	87.6&	78.9& 71.0&46.9&	93.3&	41.3&	26.7&	30.8&	58.4&	49.5&	27.2&	22.1&	42.2&	28.7&	42.5\\
+\textit{Iter I-S} & NYU  &	91.1&	82.5&76.4&	52.5&	93.6&	50.7&	35.4&	37.3&	62.8&	54.1&	34.0&	22.2&	48.7&	33.8&	47.7\\
+\textit{2-Iter I-S} & NYU  &90.7&	84.6&	77.8& 53.9&	93.2&	51.3&	38.0&	38.7&	65.0&	56.3&	37.8&	25.9&	51.3&	36.0&	49.8\\ 
\hline
Baseline  & NYUCAD  &	
92.3&	89.0&	82.8&	61.5&	94.2&	62.7&	38.0&	48.1&	69.5&	59.3&	40.1&	25.8&	54.6&	35.3&	53.6\\
+\textit{Iter I-S}  & NYUCAD  &	
93.4&	90.5&	85.1&	63.8&	94.4&	65.1&	43.0&	54.1&	73.9&	64.7&	47.8&	30.4&	60.2&	40.6&	58.0\\
+\textit{2-Iter I-S}& NYUCAD  &	
94.2&	91.3&	86.5&	65.6&	94.4&	67.1&	45.2&	57.2&	75.5&	66.4&	50.9&	31.1&	62.5&	42.9&	59.9\\
\hline

Baseline& SUNCG-RGBD  &	
90.6&	94.5&	86.0&	71.0&	86.6&	78.4&	78.7&	53.2&	77.4&	77.3&	60.6&	76.4&	83.7&	60.0&	73.0\\
+\textit{Iter I-S} & SUNCG-RGBD  &
92.2&	95.5&	88.4&	79.3&	88.8&	82.5&	80.4&	56.8&	81.2&	79.5&	63.8&	77.9&	85.3&	67.1&	76.6\\
+\textit{2-Iter I-S }  & SUNCG-RGBD  &
93.3&	96.1&	89.9&	85.2&	90.0&	83.7&	80.8&	60.0&	83.5&	80.9&	68.6&	77.3&	86.7&	70.1&	78.8\\
\hline
\end{tabular}
}
\end{center}
\vspace{-0.2cm}
\caption{Ablation studies of the effects of \textbf{scene-instance-scene iterations} on NYU, NYUCAD and SUNCG-RGBD dataset.}
\label{tab:ablation-iteration}
\vspace{-0.3cm}
\end{table*}

%% file: tables/ablation-cost.tex
\begin{table}[t]
\begin{center}
\resizebox{0.85\columnwidth}{!}{
\begin{tabular}{|c|c|c|c|c|c|c|} 
\hline
\textbf{Method} & \textbf{SC(\%)} & \textbf{SSC(\%)} &\textbf{Params/M}\\
\hline
Sketch~\cite{Chen_2020_SketchAwareSSC} & 71.3 & 41.1  &28.0\\
Ours (BiSeNet)   &77.8& 49.8 & 30.5\\
Ours (DeepLabv3)   &78.2& 52.4 & 47.3\\
\hline
\end{tabular}
}

\end{center}
\vspace{-0.2cm}
\caption{Efficiency with different methods on NYU dataset.}
\label{tab:ablation-cost}
\vspace{-0.3cm}
\end{table}

%% file: tables/ablation-instance.tex
\begin{table}[t]
\begin{center}
\resizebox{\columnwidth}{!}{
\begin{tabular}{|c|c|c|c|} 
\hline
\textbf{Scheme} & \textbf{Dataset} & \textbf{SC-IoU(\%)} & \textbf{SSC-mIoU(\%)} \\

\hline
Baseline ($S_0$) &  NYU  & 71.0 & 42.5\\
\textit{+$S_1$ only} &  NYU & 75.4 \textcolor{blue}{(+4.4)}& 44.9 \textcolor{blue}{(+2.4)}\\
\textit{+Iter I-S} &  NYU  & 76.4 \textcolor{blue}{(+5.4)} & 47.7 \textcolor{blue}{(+5.2)}\\
+\textit{2-Iter I-S} & NYU &\textbf{77.8 \textcolor{blue}{(+6.8)}}&\textbf{49.8 \textcolor{blue}{(+7.3)}}\\
\hline
Baseline ($S_0$) &  NYUCAD  & 82.8 & 53.6\\
\textit{+$S_1$ only} &  NYUCAD & 83.3 \textcolor{blue}{(+0.5)}& 54.2 \textcolor{blue}{(+0.6)}\\
\textit{+Iter I-S} &  NYUCAD & 85.1 \textcolor{blue}{(+2.3)}& 58.0 \textcolor{blue}{ (+4.4)}\\
+\textit{2-Iter I-S} & NYUCAD & \textbf{86.5 \textcolor{blue}{(+3.7)}}&\textbf{59.9 \textcolor{blue}{(+6.3)}}\\
\hline
Baseline ($S_0$) &  SUNCG-RGBD  & 86.0 & 73.0\\
\textit{+$S_1$ only} &  SUNCG-RGBD & 87.8 \textcolor{blue}{(+1.8)}& 74.5 \textcolor{blue}{(+1.5)} \\
\textit{+Iter I-S} &  SUNCG-RGBD  & 88.4 \textcolor{blue}{(+2.4)} & 76.6 \textcolor{blue}{(+3.6)}\\
+\textit{2-Iter I-S} & SUNCG-RGBD & \textbf{89.9 \textcolor{blue}{(+3.9)}}&\textbf{78.8 \textcolor{blue}{(+5.8)}}\\
\hline
\end{tabular}
}
\end{center}
\vspace{-0.2cm}
\caption{Ablation studies of the effects of \textbf{instance completion}.}
\label{tab:ablation-instance}
\vspace{-0.3cm}
\end{table}

%% file: tables/ablation-order.tex
\begin{table}[t]
\begin{center}
\resizebox{\columnwidth}{!}{
\begin{tabular}{|c|c|c|c|c|c|c|} 
\hline
\textbf{Order} & \textbf{Dataset} &\textbf{Rec.} &\textbf{mAP}
&\textbf{Shape} & \textbf{SC} & \textbf{SSC} \\
\hline
Baseline ($S_0$) &  NYU  &-&-& -& 71.0 & 42.5\\
\textit{$I_1$-$S_1$}  &  NYU   &61.7& 25.9 & 25.9 & 75.5& 46.6\\
\textit{$S_0$-$I_1$-$S_1$} &  NYU &\textbf{68.8}&\textbf{34.6} &\textbf{35.1}&\textbf{76.4} & \textbf{47.7}\\
\hline
Baseline ($S_0$) &  NYUCAD  &-&- &-&82.8 & 53.6\\
\textit{$I_{1}$-$S_{1}$} &  NYUCAD &75.0 &41.1 &30.1  & 83.1& 55.5\\
\textit{$S_0$-$I_{1}$-$S_{1}$} &  NYUCAD &\textbf{76.8} &\textbf{44.5}& \textbf{43.3} & \textbf{85.1}& \textbf{58.0}\\
\hline
Baseline ($S_{0}$) &  SUNCG-RGBD  &-&-&-& 86.0 & 73.0\\
\textit{$I_{1}$-$S_{1}$} &  SUNCG-RGBD &77.6 &65.0 &41.7 & 87.5 &75.8\\
\textit{$S_{0}$-$I_{1}$-$S_{1}$} &  SUNCG-RGBD&\textbf{81.2} &\textbf{70.4} &\textbf{50.1}  &\textbf{88.4}& \textbf{76.6}\\
\hline
\end{tabular}
}
\end{center}
\vspace{-0.2cm}
\caption{Ablation studies of effects of \textbf{initial scene completion}, where Rec. (recall) and mAP (IoU=0.25) denote the instance detection performance. Shape means the shape completion results.}
\label{tab:ablation-order}
\vspace{-0.6cm}
\end{table}